%% file: main.tex
\documentclass[10pt,twocolumn,letterpaper]{article}

\usepackage{cvpr_paper}
\usepackage{times}
\usepackage{epsfig}
\usepackage{graphicx}
\usepackage{amsmath}
\usepackage{amssymb}
\usepackage{multirow}
\usepackage{pifont}
\usepackage{makecell}

\usepackage[pagebackref=true,breaklinks=true,letterpaper=true,colorlinks,bookmarks=false]{hyperref}

\cvprfinalcopy 

\def\cvprPaperID{2659} 

\graphicspath{{figures/},{figures/qual_results/},{figures/tu-berlin-confusing/}}
\DeclareGraphicsExtensions{.pdf,.png}

\newcommand{\viz}{\textit{viz.}}
\newcommand{\fig}[1]{Figure \ref{#1}}
\newcommand{\eq}[1]{eqn. (\ref{#1})}

\newcommand{\tab}[1]{Table \ref{#1}}

\newcommand{\cmark}{\Huge\textcolor{green}{\ding{51}}}%
\newcommand{\xmark}{\Huge\textcolor{red}{\ding{55}}}%
\newcommand{\myparagraph}[1]{\vspace{4pt}\noindent{\bf #1}}

\begin{document}

\title{Semantically Tied Paired Cycle Consistency for\\Zero-Shot Sketch-based Image Retrieval}
\author{Anjan Dutta\\
Computer Vision Center\\
Autonomous University of Barcelona\\
{\tt\small adutta@cvc.uab.es}
\and
Zeynep Akata\\
Amsterdam Machine Learning Lab\\
University of Amsterdam\\
{\tt\small z.akata@uva.nl}
}

\maketitle

\input{tex/abs.tex}

\input{tex/intro.tex}
\input{tex/related.tex}

\input{tex/method.tex}

\input{tex/expt.tex}

\input{tex/concl.tex}

\input{tex/ack.tex}

{\small
\bibliographystyle{ieee}
\bibliography{bib/bibliography}
}

\end{document}

%% file: tex/abs.tex
\begin{abstract}
Zero-shot sketch-based image retrieval (SBIR) is an emerging task in computer vision, allowing to retrieve natural images relevant to sketch queries that might not been seen in the training phase. Existing works either require aligned sketch-image pairs or inefficient memory fusion layer for mapping the visual information to a semantic space. In this work, we propose a semantically aligned paired cycle-consistent generative (SEM-PCYC) model for zero-shot SBIR, where each branch maps the visual information to a common semantic space via an adversarial training. Each of these branches maintains a cycle consistency that only requires supervision at category levels, and avoids the need of highly-priced aligned sketch-image pairs. A classification criteria on the generators' outputs ensures the visual to semantic space mapping to be discriminating. Furthermore, we propose to combine textual and hierarchical side information via a feature selection auto-encoder that selects discriminating side information within a same end-to-end model. Our results demonstrate a significant boost in zero-shot SBIR performance over the state-of-the-art on the challenging Sketchy and TU-Berlin datasets.
\end{abstract}%

%% file: tex/intro.tex
\section{Introduction}
\label{sec:intro}
Matching natural images with free-hand sketches, \ie \emph{sketch-based image retrieval} (SBIR)~\cite{Yu2015,Yu2016,Liu2017DSH,Pang2017FGSBIR,Song2017SpatSemAttn,Shen2018ZSIH,Zhang2018GDH,Chen2018DeepSB3DSR,Yelamarthi2018ZSBIR} has received a lot of attention. Since sketches can effectively express shape, pose and fine-grained details of the target images, SBIR serves a favorable scenario complementary to the conventional text-image cross-modal retrieval or the classical content based image retrieval protocol. This is also because in some situations it may be hard to provide a textual description or a suitable image of the desired query, whereas, an user can easily draw a sketch of the desired object spontaneously on a touch screen.

\begin{figure}
\centering
\includegraphics[width=\columnwidth]{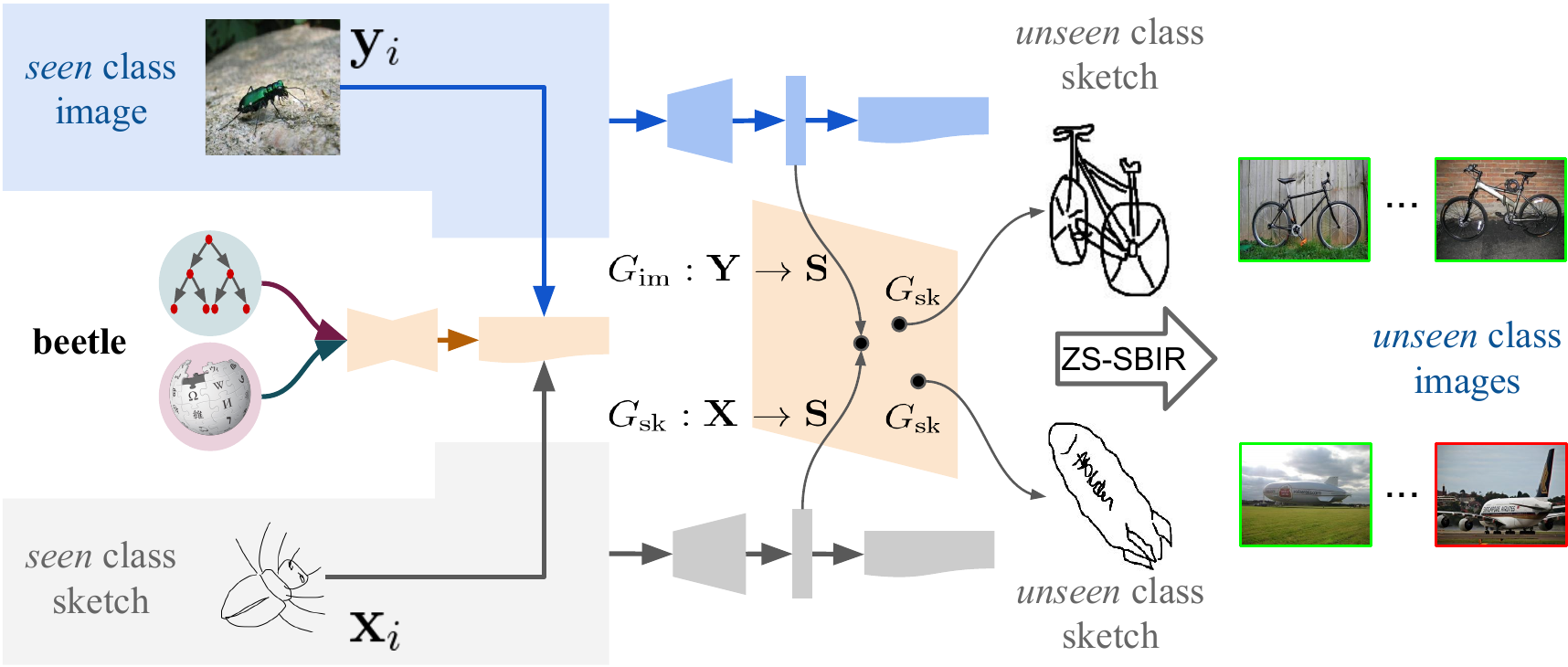}
\caption{The proposed SEM-PCYC model learns to map visual information from sketch and image to a semantic space through an adversarial training based on the \emph{seen} classes. During the testing phase the learned mappings are used for generating embeddings on the \emph{unseen} classes for zero-shot SBIR.}
\label{fig:sem-pcyc-teaser}
\end{figure}

As the visual information from all the classes gets explored by the system during training, with overlapping training and test classes, existing SBIR methods perform well~\cite{Zhang2018GDH}. Since in practice there is no guarantee that the training data would include all possible queries, a more realistic setting is \emph{zero-shot} SBIR~\cite{Shen2018ZSIH,Yelamarthi2018ZSBIR} which combines zero-shot learning (ZSL)~\cite{Lampert2014ZSL,Xian2018ZSLGBU} and SBIR as a single task, where the aim is an accurate class prediction and a competent retrieval performance. However, zero-shot SBIR is extremely challenging as it simultaneously deals with a significant domain gap, intra-class variability and limited knowledge about the \emph{unseen} classes. 

One of the major shortcomings of the prior work on ZS-SBIR is that sketch-image is retrieved after learning a mapping from an input sketch to an output image using a training set of labelled \emph{aligned} pairs~\cite{Yelamarthi2018ZSBIR}. The supervision of paired correspondence is to enhance the correlation of multi-modal data (here, sketch-image) so that learning can be guided by semantics. However, for many realistic scenarios, obtaining paired (aligned) training data is either unavailable or very expensive. Furthermore, often a joint representation of two or more modalities is obtained by using a memory fusion layer~\cite{Shen2018ZSIH}, such as, tensor fusion~\cite{Hu2017TFN}, bilinear pooling~\cite{Yu2017MFBP} etc. These fusion layers are often expensive in terms of memory~\cite{Yu2017MFBP}, and extracting useful information from this high dimensional space could result in information loss~\cite{Yu2018MHBN}.

To alleviate these shortcomings, we propose a semantically aligned paired cycle consistent generative (SEM-PCYC) model for zero-shot SBIR task, where each branch either maps sketch or image features to a common semantic space via an adversarial training. These two branches dealing with two different modalities (sketch and image) constitute an essential component for solving SBIR task. The cycle consistency constraint on each branch guarantees the mapping of sketch or image modality to a common semantic space and their translation back to the original modality, which further avoids the necessity of aligned sketch-image pairs. Imposing a classification loss on the semantically aligned outputs from the sketch and image space enforces the generated features in the semantic space to be discriminative which is very crucial for effective zero-shot SBIR. Furthermore, inspired by the previous works on label embedding~\cite{Akata2015OutputEmbedding}, we propose to combine side information from text-based and hierarchical models via a feature selection auto-encoder~\cite{Wang2017FSAE} which selects discriminating side information based on intra and inter class covariance.


The main contributions of the paper are: (1) We propose the SEM-PCYC model for zero-shot SBIR task, that maps sketch and image features to a common semantic space with the help of adversarial training. The cycle consistency constraint on each branch of the SEM-PCYC model facilitates bypassing the requirement of aligned sketch image pairs. (2) Within a same end-to-end framework, we combine different side information via a feature selection guided auto-encoder which effectively choose side information that minimizes intra-class variance and maximizes inter-class variance. (3) We evaluate our model on two datasets (Sketchy and TU-Berlin) with varying difficulties and sizes, and provide an experimental comparison with latest models available for the same task, which further shows that our proposed model consistently improves the state-of-the-art results of zero-shot SBIR on both datasets.

%% file: tex/related.tex
\section{Related Work}
\label{sec:related}
As our work belongs at the verge of sketch-based image retrieval and zero-shot learning task, we briefly review the relevant literature from both the fields.

\myparagraph{Sketch Based Image Retrieval (SBIR).}
Attempts for solving SBIR task mostly focus on bridging the domain gap between sketch and image, which can roughly be grouped in \emph{hand-crafted} and \emph{cross-domain deep learning-based} methods~\cite{Liu2017DSH}. Hand-crafted methods mostly work by extracting the edge map from natural image and then matching them with sketch using a Bag-of-Words model on top of some specifically designed SBIR features, \viz, gradient field HOG~\cite{Hu2013}, histogram of oriented edges~\cite{Saavedra2014SoftComp}, learned key shapes~\cite{Saavedra2015LKS} etc. However, the difficulty of reducing domain gap remained unresolved as it is extremely challenging to match edge maps with unaligned hand drawn sketch. This domain shift issue is further addressed by neural network models where domain transferable features from sketch to image are learned in an end-to-end manner. Majority of such models use variant of siamese networks~\cite{Qi2016SBIRSiamese,Sangkloy2016,Yu2016,Song2017FineGrained} that are suitable for cross-modal retrieval. These frameworks either use generic ranking losses, \viz, contrastive loss~\cite{Chopra2005}, triplet ranking loss~\cite{Sangkloy2016} or more sophisticated HOLEF based loss~\cite{Song2017SpatSemAttn}) for the same. Further to these discriminative losses, Pang~\etal~\cite{Pang2017FGSBIR} introduced a discriminative-generative hybrid model for preserving all the domain invariant information useful for reducing the domain gap between sketch and image. Alternatively, some other works focus on learning cross-modal hash code for category level SBIR within an end-to-end deep model~\cite{Liu2017DSH,Zhang2018GDH}. In contrast, we propose a paired cycle consistent generative model where each branch either maps sketch or image features to a common semantic space via adversarial training, which we found to be effective for reducing the domain gap between sketch and image.

\myparagraph{Zero-Shot Learning (ZSL).}
Zero-shot learning in computer vision refers to recognizing objects whose instances are not seen during the training phase; a comprehensive and detailed survey on ZSL is available in~\cite{Xian2018ZSLGBU}. Early works on ZSL~\cite{Lampert2014ZSL,Jayaraman2014ZSR,Changpinyo2016ZSL,Al-Halah2016ZSL} make use of attributes within a two-stage approach to infer the label of an image that belong to the \emph{unseen} classes. However, the recent works~\cite{Frome2013Devise,Romera-Paredes2015ESA,Akata2015OutputEmbedding,Akata2016LabelEmbedding,Kodirov2017SAE} directly learn a mapping from image feature space to a semantic space. Many other ZSL approaches learn non-linear multi-modal embedding~\cite{Socher2013ZSLCrossModalT,Akata2016LabelEmbedding,Xian2016ZSLLatentEmbedding,Changpinyo2017ZSL,Zhang2017ZSLDeepEmbedding}, where most of the methods focus to learn a non-linear mapping from the image space to the semantic space. Mapping both image and semantic features into another common intermediate space is another direction that ZSL approaches adapt~\cite{Zhang2015ZSLSemSim,Fu2015ZSOR,Zhang2016ZSLJointLatentSim,Akata2016ZSLSS,Long2017ZSL}. Although, most of the deep neural network models in this domain are trained using a discriminative loss function, a few generative models also exist~\cite{Wang2018ZSL,Xian2018ZSL,Chen2018ZSVR} that are used as a data augmentation mechanism. In ZSL, some form of side information is required, so that the knowledge learned from \emph{seen} classes gets transferred to \emph{unseen} classes. One popular form of side information is attributes~\cite{Lampert2014ZSL} that, however, require costly expert annotation. Thus, there has been a large group of studies~\cite{Mensink2014COSTA,Akata2015OutputEmbedding,Xian2016ZSLLatentEmbedding,Reed2016LDR,Qiao2016LiM,Ding2017} which utilize other auxiliary information, such as, text-based~\cite{Mikolov2013a} or hierarchical model~\cite{Miller1995WN} for label embedding. In this work, we address zero-shot cross-modal (sketch to image) retrieval, for that, motivated by~\cite{Akata2015OutputEmbedding}, we effectively combine different side information within an end-to-end framework, and map visual information to the semantic space through an adversarial training.

\myparagraph{Zero-Shot Sketch-based Image Retrieval (ZS-SBIR).}
Shen~\etal~\cite{Shen2018ZSIH} first combined zero-shot learning and sketch based image retrieval, and proposed a generative cross-modal hashing scheme for solving the zero-shot SBIR task, where they used a graph convolution network for aligning the sketch and image in the semantic space. Inspired by them, Yelamarthi~\etal~\cite{Yelamarthi2018ZSBIR} proposed two similar autoencoder-based generative models for zero-shot SBIR, where they have used the aligned pairs of sketch and image for learning the semantics between them. In contrast, we propose a paired cycle-consistent generative model where each branch maps the visual information from sketch or image to a semantic space through an adversarial training with a common discriminator. The cycle consistency constraint on each branch allows supervision only at category level, and avoids the need of aligned sketch-image pairs.

%% file: tex/method.tex
\begin{figure*}[t]
\centering
\includegraphics[width=\textwidth]{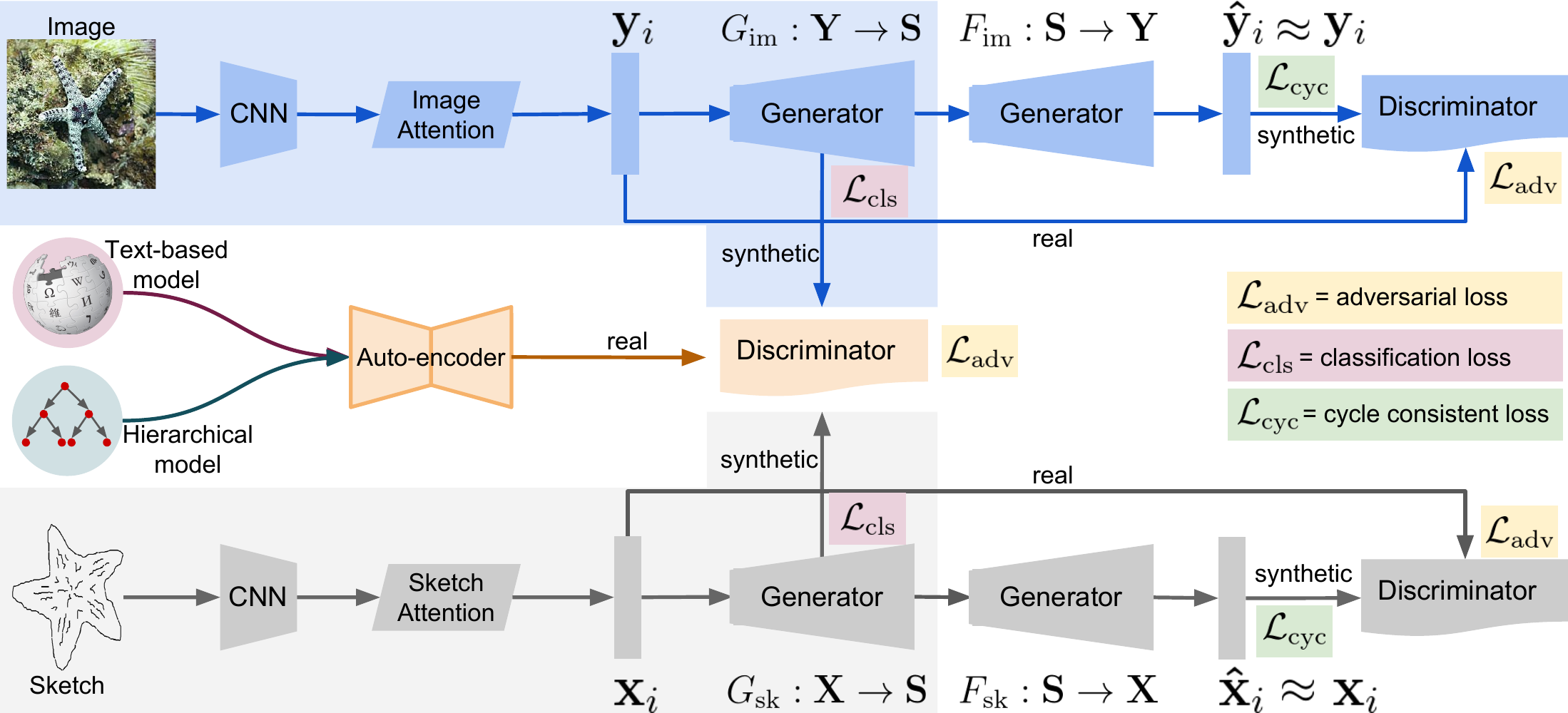}
\caption{The deep network structure of SEM-PCYC. The sketch (in light gray) and image cycle consistent networks (in light blue) respectively map the sketch and image to the semantic space and then the original input space. An auto-encoder (light orange) combines the semantic information based on text and hierarchical model, and produces a compressed semantic representation which acts as a true example to the discriminator. During the test phase only the learned sketch (light gray region) and image (light blue region) encoders to the semantic space are used for generating embeddings on the \emph{unseen} classes for zero-shot SBIR. (best viewed in color)}
\label{fig:sem-pcyc}
\end{figure*}

\section{SEM-PCYC Model}
\label{sec:method}
In this work, we propose the semantically aligned paired cycle consistent generative (SEM-PCYC) model for zero-shot sketch-based image retrieval. The sketch and image data from the \emph{seen} categories are only used for training the underlying model. Our SEM-PCYC model encodes and matches the sketch and image categories that remain \emph{unseen} during the training phase. The overall pipeline of our end-to-end deep architecture is shown in~\fig{fig:sem-pcyc}.

Let $\mathcal{D}^s=\{\mathbf{X}^s, \mathbf{Y}^s\}$ be a collection of sketch and image data from the \emph{seen} categories $\mathcal{C}^s$ that contain sketch images $\mathbf{X}^s=\{\mathbf{x}_i^s\}_{i=1}^N$ as well as natural images $\mathbf{Y}^s=\{\mathbf{y}_i^s\}_{i=1}^N$ for training, where $N$ is the total number of sketch and image pairs that are not necessarily aligned. Without loss of generality, it can be assumed that sketch and image having the same index, say, $i$, share the same category label. Let $\mathbf{S}^s=\{\mathbf{s}_i^s\}_{i=1}^{N}$ be the set of side information useful for transferring the supervised knowledge to the \emph{unseen} classes, which is an usual practice in ZSL methods. The main aim of our model is to learn two deep functions $G_\text{sk}(\cdot)$ and $G_\text{im}(\cdot)$ respectively for sketch and image for mapping them to a common semantic space where the learned knowledge can be applied to the \emph{unseen} classes as well. Given a set of sketch-image data $\mathcal{D}^u=\{\mathbf{X}^u, \mathbf{Y}^u\}$ from the \emph{unseen} categories $\mathcal{C}^u$ for test, the proposed deep functions $G_\text{sk}:\mathbb{R}^d\rightarrow\mathbb{R}^M$, $G_\text{im}:\mathbb{R}^d\rightarrow\mathbb{R}^M$ ($d$ is the dimension of the original data and $M$ is the targeted dimension of the common representation) map the sketch and natural image to a common semantic space where the retrieval is performed. Since the method considers SBIR in zero-shot setting, it is evident that the \emph{seen} and \emph{unseen} categories remain exclusive, \ie $\mathcal{C}^s\cap\mathcal{C}^u=\varnothing$.

\subsection{Paired Cycle Consistent Generative Model}
\label{ssec:cyc-con-gen-model}

For having the flexibility to handle sketch and image individually, \ie even when they are not aligned sketch-image pairs, during training $G_\text{sk}$ and $G_\text{im}$, we propose a cycle consistent generative model whose each branch is semantically aligned with a common discriminator. The cycle consistency constraint on each branch of the model ensures the mapping of sketch or image modality to a common semantic space, and their translation back to the original modality, which only requires supervision at category level. Imposing a classification loss on the output of $G_\text{sk}$ and $G_\text{im}$ allows generating highly discriminative features.

Our main goal is to learn two mappings $G_\text{sk}$ and $G_\text{im}$ that can respectively translate the unaligned sketch and natural image to a common semantic space. Zhu~\etal~\cite{Zhu2017CycleGAN} pointed out about the existence of underlying intrinsic relationship between modalities and domains, for example, sketch or image of same object category have the same semantic meaning, and possess that relationship. Even though, we lack visual supervision as we do not have access to aligned pairs, we can exploit semantic supervision at category levels. We train a mapping $G_\text{sk}:\mathbf{X}\rightarrow\mathbf{S}$
so that $\hat{\mathbf{s}}_i=G_\text{sk}(\mathbf{x}_i)$, where $\mathbf{s}_i\in\mathbf{S}$ is the corresponding side information and is indistinguishable from $\hat{\mathbf{s}}_i$ via an adversarial training that classifies $\hat{\mathbf{s}}_i$ different from $\mathbf{s}_i$. The optimal $G_\text{sk}$ thereby translates the modality $\mathbf{X}$ into a modality $\hat{\mathbf{S}}$ which is identically distributed to $\mathbf{S}$. Similarly, another function $G_\text{im}:\mathbf{Y}\rightarrow\mathbf{S}$ can be trained via the same discriminator such that $\hat{\mathbf{s}}_i=G_\text{im}(\mathbf{y}_i)$.

\myparagraph{Adversarial Loss.}
As shown in~\fig{fig:sem-pcyc}, for mapping the sketch and image representation to a common semantic space, we introduce four generators $G_\text{sk}:\mathbf{X}\rightarrow\mathbf{S}$, $G_\text{im}:\mathbf{Y}\rightarrow\mathbf{S}$, $F_\text{sk}:\mathbf{S}\rightarrow\mathbf{X}$ and $F_\text{im}:\mathbf{S}\rightarrow\mathbf{Y}$. In addition, we bring in three adversarial discriminators: $D_\text{se}(\cdot)$, $D_\text{sk}(\cdot)$ and $D_\text{im}(\cdot)$, where $D_\text{se}$ discriminates among original side information $\{\mathbf{s}\}$, sketch transformed to side information $\{G_\text{sk}(\mathbf{x})\}$ and image transformed to side information $\{G_\text{im}(\mathbf{y})\}$; likewise $D_\text{sk}$ discriminates between original sketch representation $\{\mathbf{x}\}$ and side information transformed to sketch representation $\{F_\text{sk}(\mathbf{s})\}$; in a similar way $D_\text{im}$ distinguishes between $\{\mathbf{y}\}$ and $\{F_\text{im}(\mathbf{s})\}$. For the generators $G_\text{sk}$, $G_\text{im}$ and their common discriminator $D_\text{se}$, the objective is as follows:
\begin{align}
&\mathcal{L}_\text{adv}(G_\text{sk}, G_\text{im}, D_\text{se}, \mathbf{x}, \mathbf{y}, \mathbf{s})=2\times\mathbb{E}\left[\log D_\text{se}(\mathbf{s})\right]\\
&+\mathbb{E}\left[\log(1- D_\text{se}(G_\text{sk}(\mathbf{x})))\right]+\mathbb{E}\left[\log(1- D_\text{se}(G_\text{im}(\mathbf{y})))\right] \nonumber
\label{eqn:adv_sem}
\end{align}
where $G_\text{sk}$ and $G_\text{im}$ generate side information similar to the ones in $\mathbf{S}$ while $D_\text{se}$ distinguishes between the generated and original side information. Here, $G_\text{sk}$ and $G_\text{im}$ minimize the objective against an opponent $D_\text{se}$ that tries to maximize it, \ie $\min_{G_\text{sk}, G_\text{im}}\max_{D_\text{se}}\mathcal{L}_\text{adv}(G_\text{sk}, G_\text{im}, D_\text{se}, \mathbf{x}, \mathbf{y}, \mathbf{s})$. In a similar way, for the generator $F_\text{sk}$ and its discriminator $D_\text{sk}$, the objective is:
\begin{equation}
\begin{split}
\mathcal{L}_\text{adv}(F_\text{sk}, D_\text{sk}, \mathbf{x}, \mathbf{s})=&\text{ }\mathbb{E}\left[\log D_\text{sk}(\mathbf{x})\right]\\
&+\mathbb{E}\left[\log(1- D_\text{sk}(F_\text{sk}(\mathbf{s})))\right]
\end{split}
\label{eqn:adv_sk}
\end{equation}
$F_\text{sk}$ minimizes the objective and its adversary $D_\text{sk}$ intends to maximize it, \ie $\min_{F_\text{sk}}\max_{D_\text{sk}}\mathcal{L}_\text{adv}(F_\text{sk}, D_\text{sk}, \mathbf{x}, \mathbf{s})$. Similarly, another adversarial loss is introduced for the mapping $F_\text{im}$ and its discriminator $D_\text{im}$, \ie, $\min_{F_\text{im}}\max_{D_\text{im}}\mathcal{L}_\text{adv}(F_\text{im}, D_\text{im}, \mathbf{y}, \mathbf{s})$.

\myparagraph{Cycle Consistency Loss.}
The adversarial mechanism effectively reduces the domain or modality gap, however, it is not guaranteed that an input $\mathbf{x}_i$ and an output $\mathbf{s}_i$ are matched well. To this end, we impose cycle consistency~\cite{Zhu2017CycleGAN}. When we map the feature of a sketch of an object to the corresponding semantic space, and then further translate it back from the semantic space to the sketch feature space, we should reach back to the original sketch feature. This cycle consistency loss also assists in learning mappings across domains where paired or aligned examples are not available. Specifically, if we have a function $G_\text{sk}:\mathbf{X}\rightarrow\mathbf{S}$ and another mapping $F_\text{sk}:\mathbf{S}\rightarrow\mathbf{X}$, then both $G_\text{sk}$ and $F_\text{sk}$ are reverse of each other, and hence form a one-to-one correspondence or bijective mapping.
\begin{equation}
\begin{split}
\mathcal{L}_\text{cyc}(G_\text{sk}, F_\text{sk})=&\text{ }\mathbb{E}\left[\Vert F_\text{sk}(G_\text{sk}(\mathbf{x}))-\mathbf{x} \Vert_{1}\right]\\
&+\mathbb{E}\left[\Vert G_\text{sk}(F_\text{sk}(\mathbf{s}))-\mathbf{s} \Vert_{1}\right]
\end{split}
\end{equation}
Similarly, a cycle consistency loss is imposed for the mappings $G_\text{im}:\mathbf{Y}\rightarrow\mathbf{S}$ and $F_\text{im}:\mathbf{S}\rightarrow\mathbf{Y}$: $\mathcal{L}_\text{cyc}(G_\text{im}, F_\text{im})$. These consistent loss functions also behave as a regularizer to the adversarial training to assure that the learned function maps a specific input $\mathbf{x}_i$ to a desired output $\mathbf{s}_i$.

\myparagraph{Classification Loss.}
On the other hand, adversarial training and cycle-consistency constraints do not explicitly ensure whether the generated features by the mappings $G_\text{sk}$ and $G_\text{im}$ are class discriminative, \ie a requirement for the zero-shot sketch-based image retrieval task. We conjecture that this issue can be alleviated by introducing a discriminative classifier pre-trained on the input data. At this end we  minimize a classification loss over the generated features.
\begin{equation}
\mathcal{L}_\text{cls}(G_\text{sk})=-\mathbb{E}\left[\log P(c|G_\text{sk}(\mathbf{x});\theta) \right]
\end{equation}
where $c$ is the category label of $\mathbf{x}$. Similarly, a classification loss $\mathcal{L}_\text{cls}(G_\text{im})$ is also imposed on the generator $G_\text{im}$.

\subsection{Selection of Side Information}
\label{ssec:autoenc}
Motivated by attribute selection for zero-shot learning~\cite{Guo2018ZSL}, indicating that a subset of discriminative attributes are more effective than the whole set of attributes for ZSL, we incorporate a joint learning framework integrating an auto-encoder to select side information. Let $\mathbf{s}\in\mathbb{R}^{k}$ be the side information with $k$ as the original dimension. The loss function is: 
\begin{equation}
\mathcal{L}_\text{aenc}(f,g)=\Vert \mathbf{s}-g(f(\mathbf{s}))\Vert_{F}^{2}+\lambda \Vert W_1 \Vert_{2,1}
\label{eqn:aenc_loss}
\end{equation}
where $f(\mathbf{s})=\sigma(W_1\mathbf{s}+b_1)$, $g(f(\mathbf{s}))=\sigma(W_2f(\mathbf{s})+b_2)$, with $W_1\in\mathbb{R}^{k\times m}$, $W_2\in\mathbb{R}^{m\times k}$ and $b_1$, $b_2$ respectively as the weights and biases for the function $f$ and $g$. Selecting side information reduces the dimensionality of embeddings, which further improves retrieval time. Therefore, the training objective of our model:
\begin{align}
&\mathcal{L}(G_\text{sk}, G_\text{im}, F_\text{sk}, F_\text{im}, D_\text{se}, D_\text{sk}, D_\text{im}, f, g, \mathbf{x}, \mathbf{y}, \mathbf{s})\nonumber\\
&= \mathcal{L}_\text{adv}(G_\text{sk}, G_\text{im}, D_\text{se}, \mathbf{x}, \mathbf{y}, \mathbf{s})+\mathcal{L}_\text{adv}(F_\text{sk}, D_\text{sk}, \mathbf{x}, \mathbf{s})\\
&+\mathcal{L}_\text{adv}(F_\text{im}, D_\text{im}, \mathbf{y}, \mathbf{s})+\mathcal{L}_\text{cyc}(G_\text{sk}, F_\text{sk})+\mathcal{L}_\text{cyc}(G_\text{im}, F_\text{im})\nonumber\\
&+\mathcal{L}_\text{cls}(G_\text{sk})+\mathcal{L}_\text{cls}(G_\text{im})+\mathcal{L}_\text{aenc}(f,g)\nonumber
\label{eqn:comb-loss}
\end{align}

For obtaining the initial side information, we combine a text-based and a hierarchical model, which are complementary and robust~\cite{Akata2015OutputEmbedding}. Below, we provide a description of our text-based and hierarchical models for side information.

\myparagraph{Text-based Model.}
We use two different text-based side information. (1) Word2Vec~\cite{Mikolov2013} is a two layered neural network that are trained to reconstruct linguistic contexts of words. During training, it takes a large corpus of text and creates a vector space of several hundred dimensions, with each unique word being assigned to a corresponding vector in that space. The model can be trained with a hierarchical softmax with either skip-gram or continuous bag-of-words formulation for target prediction. (2) GloVe~\cite{Pennington2014GloVe} considers global word-word co-occurrence statistics that frequently appear in a corpus. Intuitively, co-occurrence statistics encode important semantic information. The objective is to learn word vectors such that their dot product equals to the probability of their co-occurrence.

\myparagraph{Hierarchical Model.}
Semantic similarity between words can also be approximated by measuring their distance in a large onthology such as WordNet\footnote{\url{https://wordnet.princeton.edu}} of $\approx 100,000$ words in English. One can measure similarity using techniques such as path similarity and Jiang-Conrath~\cite{Jiang1997SemSim}. For a set $\mathbb{S}$ of nodes in a dictionary $\mathbb{D}$, similarities between every class $c$ and all the other nodes in $\mathbb{S}$ determine the entries of the class embedding vector~\cite{Akata2015OutputEmbedding}. $\mathbb{S}$ considers all the nodes on the path from each node in $\mathbb{D}$ to its highest level ancestor. The database of WordNet contains most of the classes of the Sketchy~\cite{Sangkloy2016} and Tu-Berlin~\cite{Eitz2012TUBerlin} datasets. Few exceptions are: \emph{jack-o-lantern} which we replaced with \emph{lantern} that appears higher in the hierarchy, similarly \emph{human skeleton} with \emph{skeleton}, and \emph{octopus} with \emph{octopods} etc. $|\mathbb{S}|$ for Sketchy and TU-Berlin datasets are respectively $354$ and $664$. 


%% file: tex/expt.tex
\section{Experiments}
\label{sec:expt}
\myparagraph{Datasets.} We experimentally validate our model on two popular SBIR benchmarks: Sketchy~\cite{Sangkloy2016} and TU-Berlin~\cite{Eitz2012TUBerlin}, together with the extended images from~\cite{Liu2017DSH}.

The Sketchy Dataset~\cite{Sangkloy2016} (Extended) is a large collection of sketch-photo pairs. The dataset consists of images from $125$ different classes, with $100$ photos each. Sketch images of the objects that appear in these $12,500$ images are collected via crowd sourcing, which resulted in $75,471$ sketches. This dataset also contains a fine grained correspondence (aligned) between particular photos and sketches as well as various data augmentations for deep learning based methods. Liu~\etal~\cite{Liu2017DSH} extended the dataset by adding $60,502$ photos yielding in total $73,002$ images. We randomly pick $25$ classes of sketches and images as the \emph{unseen} test set for the zero-shot SBIR, and the data from remaining $100$ \emph{seen} classes are used for training.

The TU-Berlin Dataset~\cite{Eitz2012TUBerlin} (Extended) contains $250$ categories with a total of $20,000$ sketches extended by~\cite{Liu2017DSH} with natural images corresponding to the sketch classes with a total size of $204,489$. $30$ classes of sketches and images are randomly chosen to respectively form the query set and the retrieval gallery. The remaining $220$ classes are utilized for training. We follow Shen~\etal~\cite{Shen2018ZSIH} and select classes with at least $400$ images in the test set.


{
\setlength{\tabcolsep}{6pt}
\renewcommand{\arraystretch}{0.8} 
\begin{table*}[!t]
\centering
\resizebox{\textwidth}{!}{
\begin{tabular}{l l cccc|cccc}
  &  & \multicolumn{4}{c|}{\textbf{Sketchy (Extended)}} & \multicolumn{4}{c}{\textbf{TU-Berlin (Extended)}} \\
 & \textbf{Method} & \textbf{mAP} & \textbf{Precision} & \textbf{Feature} & \textbf{Retrieval} & \textbf{mAP} & \textbf{Precision} & \textbf{Feature} & \textbf{Retrieval} \\
  &  & \textbf{@all} & \textbf{@100} & \textbf{Dimension} & \textbf{Time (s)} & \textbf{@all} & \textbf{@100} & \textbf{Dimension} & \textbf{Time (s)} \\
\hline
\multirow{7}{*}{SBIR} & Softmax Baseline & $0.114$ & $0.172$ & $4096$ & $3.5\times10^{-1}$ & $0.089$ & $0.143$ & $4096$ & $4.3\times10^{-1}$ \\
  & Siamese CNN~\cite{Qi2016SBIRSiamese} & $0.132$ & $0.175$ & $64$ & $5.7\times10^{-3}$ & $0.109$ & $0.141$ & $64$ & $5.9\times10^{-3}$ \\
  & SaN~\cite{Yu2016a} & $0.115$ & $0.125$ & $512$ & $4.8\times10^{-2}$ & $0.089$ & $0.108$ & $512$ & $5.5\times10^{-2}$ \\
  & GN Triplet~\cite{Sangkloy2016} & $0.204$ & $0.296$ & $1024$ & $9.1\times10^{-2}$ & $0.175$ & $0.253$ & $1024$ & $1.9\times10^{-1}$ \\
  & 3D Shape~\cite{Wang2015} & $0.067$ & $0.078$ & $64$ & $7.8\times10^{-3}$ & $0.054$ & $0.067$ & $64$ & $7.2\times10^{-3}$ \\
  & DSH (binary)~\cite{Liu2017DSH} & $0.171$ & $0.231$ & $64$ & $6.1\times10^{-5}$ & $0.129$ & $0.189$ & $64$ & $7.2\times10^{-5}$ \\
  & GDH (binary)~\cite{Zhang2018GDH} 	& $0.187$ & $0.259$ & $64$ & $7.8\times10^{-5}$ & $0.135$ & $0.212$ & $64$ & $9.6\times10^{-5}$ \\
\hline
\multirow{7}{*}{ZSL} & CMT~\cite{Socher2013ZSLCrossModalT} & $0.087$ & $0.102$ & $300$ & $2.8\times10^{-2}$ & $0.062$ & $0.078$ & $300$ & $3.3\times10^{-2}$ \\
  & DeViSE~\cite{Frome2013Devise} & $0.067$ & $0.077$ & $300$ & $3.6\times10^{-2}$ & $0.059$ & $0.071$ & $300$ & $3.2\times10^{-2}$ \\
  & SSE~\cite{Zhang2015BitScalable} & $0.116$ & $0.161$ & $100$ & $1.3\times10^{-2}$ & $0.089$ & $0.121$ & $220$ & $1.7\times10^{-2}$ \\
  & JLSE~\cite{Zhang2016ZSLJointLatentSim} 	& $0.131$ & $0.185$ & $100$ & $1.5\times10^{-2}$ & $0.109$ & $0.155$ & $220$ & $1.4\times10^{-2}$ \\
  & SAE~\cite{Kodirov2017SAE} & $0.216$ & $0.293$ & $300$ & $2.9\times10^{-2}$ & $0.167$ & $0.221$ & $300$ & $3.2\times10^{-2}$ \\
  & FRWGAN~\cite{Felix2018FRWGAN} & $0.127$ & $0.169$ & $512$ & $3.2\times10^{-2}$ & $0.110$ & $0.157$ & $512$ & $3.9\times10^{-2}$ \\
  & ZSH (binary)~\cite{Yang2016} & $0.159$ & $0.214$ & $64$  & $5.9\times10^{-5}$ & $0.141$ & $0.177$ & $64$ & $7.6\times10^{-5}$ \\
\hline
\multirow{4}{*}{Zero-Shot SBIR} & ZSIH (binary)~\cite{Shen2018ZSIH} & $0.258$ & $0.342$ & $64$ & $6.7\times10^{-5}$ & $0.223$ & $0.294$ & $64$ & $7.7\times10^{-5}$ \\
  & ZS-SBIR~\cite{Yelamarthi2018ZSBIR} & $0.196$ & $0.284$ & $1024$ & $9.6\times10^{-2}$ & $0.005$ & $0.001$ & $1024$ & $1.2\times10^{-1}$ \\
  & \textbf{SEM-PCYC} & $\mathbf{0.349}$ & $\mathbf{0.463}$ & $64$ & $1.7\times 10^{-3}$ & $\mathbf{0.297}$ & $\mathbf{0.426}$ & $64$ & $1.9\times 10^{-3}$ \\
  & \textbf{SEM-PCYC (binary)} & $\mathbf{0.344}$ & $\mathbf{0.399}$ & $64$ & $9.5 \times 10^{-5}$ & $\mathbf{0.293}$ & $\mathbf{0.392}$ & $64$ & $9.3\times 10^{-4}$ \\
\hline
\multirow{3}{*}{\makecell{Generalized\\Zero-Shot SBIR}} & ZSIH (binary)~\cite{Shen2018ZSIH} & $0.219$ & $0.296$ & $64$ & $6.7\times10^{-5}$ & $0.142$ & $0.218$ & $64$ & $7.7\times10^{-5}$ \\
 & \textbf{SEM-PCYC} & $\mathbf{0.307}$ & $\mathbf{0.364}$ & $64$ & $1.7\times 10^{-3}$ & $\mathbf{0.192}$ & $\mathbf{0.298}$ & $64$ & $2.0\times 10^{-3}$ \\
 & \textbf{SEM-PCYC (binary)} & $\mathbf{0.260}$ & $\mathbf{0.317}$ & $64$ & $9.4 \times 10^{-5}$ & $\mathbf{0.174}$ & $\mathbf{0.267}$ & $64$ & $9.3\times 10^{-4}$ \\
\end{tabular}
}
\caption{Zero-shot sketch-based image retrieval performance comparison with existing SBIR, ZSL, zero-shot SBIR and generalized zero-shot SBIR methods. Note: SBIR and ZSL methods are adapted to the Zero-Shot SBIR task, same \emph{seen} and \emph{unseen} classes are used for a fair comparison.}
\label{tab:results_z2sbir}
\end{table*}
}
\myparagraph{Implementation Details.}
We implemented the SEM-PCYC model using PyTorch~\cite{Paszke2017PyTorch} deep learning toolbox\footnote{Our code and trained models are available at: \url{https://github.com/AnjanDutta/sem-pcyc}}, which is trainable on a single TITAN Xp graphics card. We extract features from sketch and image from the VGG-$16$~\cite{Simonyan2014} network model pre-trained on ImageNet~\cite{Deng2009ImageNet} dataset (before the last pooling layer). Since in this work, we deal with single object retrieval and an object usually spans only on certain regions of a sketch or image, we apply an attention mechanism inspired by Song~\etal~\cite{Song2017SpatSemAttn} without the shortcut connection for extracting only the informative regions from sketch and image. The attended $512$-D representation is obtained by a pooling operation guided by the attention model and fully connected (fc) layer. This entire model is fine tuned on our training set ($100$ classes for Sketchy and $220$ classes for TU-Berlin). Both the generators $G_\text{sk}$ and $G_\text{im}$ are built with a fc layer followed by a ReLU non-linearity that accept $512$-D vector and output $M$-D representation, whereas, the generators $F_\text{sk}$ and $F_\text{im}$ take $M$-D features and produce $512$-D vector. Accordingly, all discriminators are designed to take the output of respective generators and produce a single dimensional output. The auto-encoder is designed by stacking two non-linear fc layers respectively as encoder and decoder for obtaining a compressed and encoded representation of dimension $M$.

While constructing the hierarchy for acquiring the class embedding, we only consider the \emph{seen} classes belong to that dataset. In this way, the WordNet hierarchy or the knowledge graph for the Sketchy and TU-Berlin datasets respectively contain $354$ and $664$ nodes. Although our method does not produce binary hash code as a final representation for matching sketch and image, for the sake of comparison with some related works, such as, ZSH~\cite{Yang2016ZSH}, ZSIH~\cite{Shen2018ZSIH}, GDH~\cite{Zhang2018GDH}, that produce hash codes, we have used the iterative quantization (ITQ)~\cite{Gong2013ITQ} algorithm to obtain the binary codes for sketch and image. We have used final representation of sketches and images from the train set to learn the optimized rotation which later used on our final representation for obtaining the binary codes.

\begin{figure}[!t]
\centering
\resizebox{\columnwidth}{!}{
\begin{tabular}{@{}c@{}c@{}c@{}c@{}c}
\includegraphics[width=0.25\columnwidth]{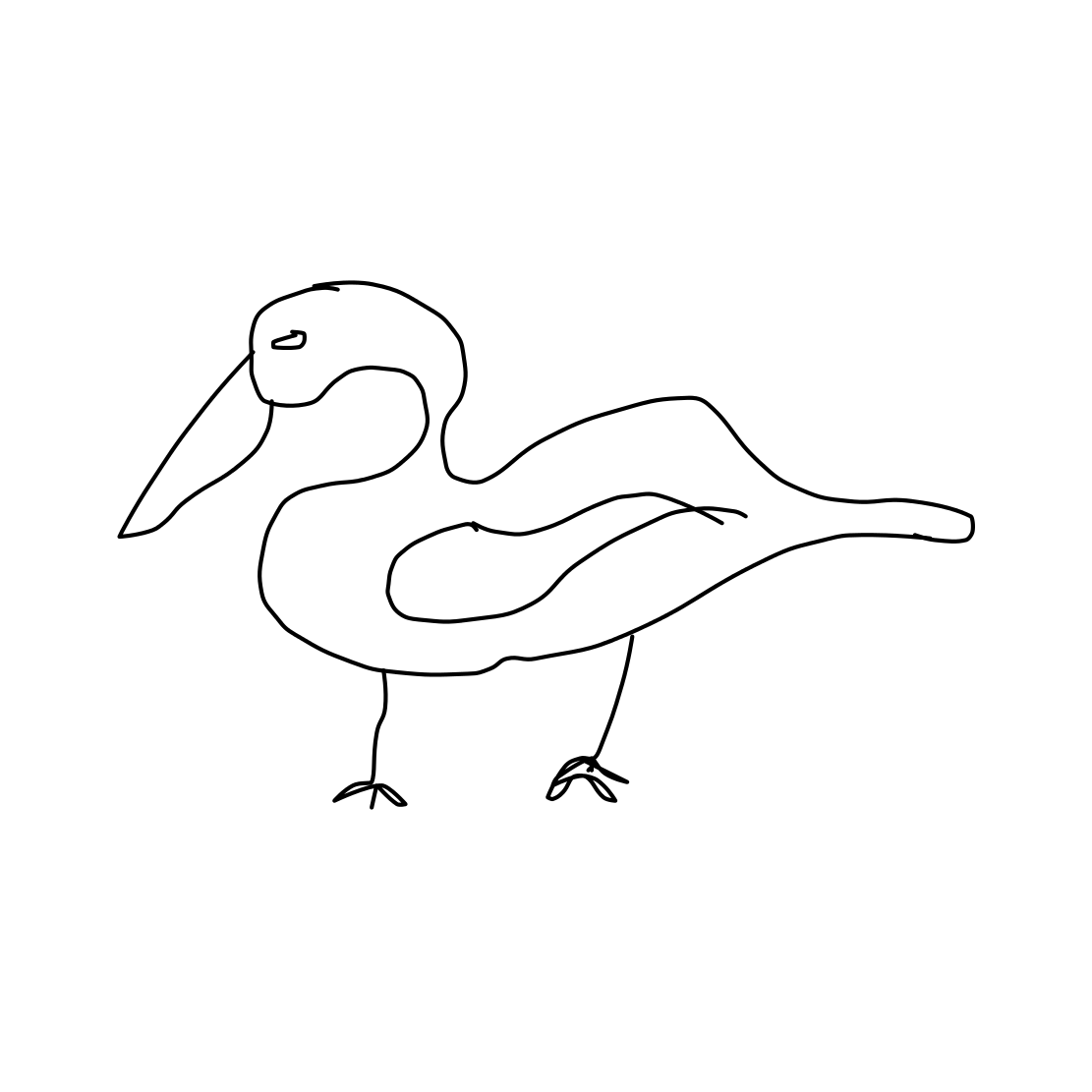} &
\includegraphics[width=0.25\columnwidth]{duck} & \includegraphics[width=0.25\columnwidth]{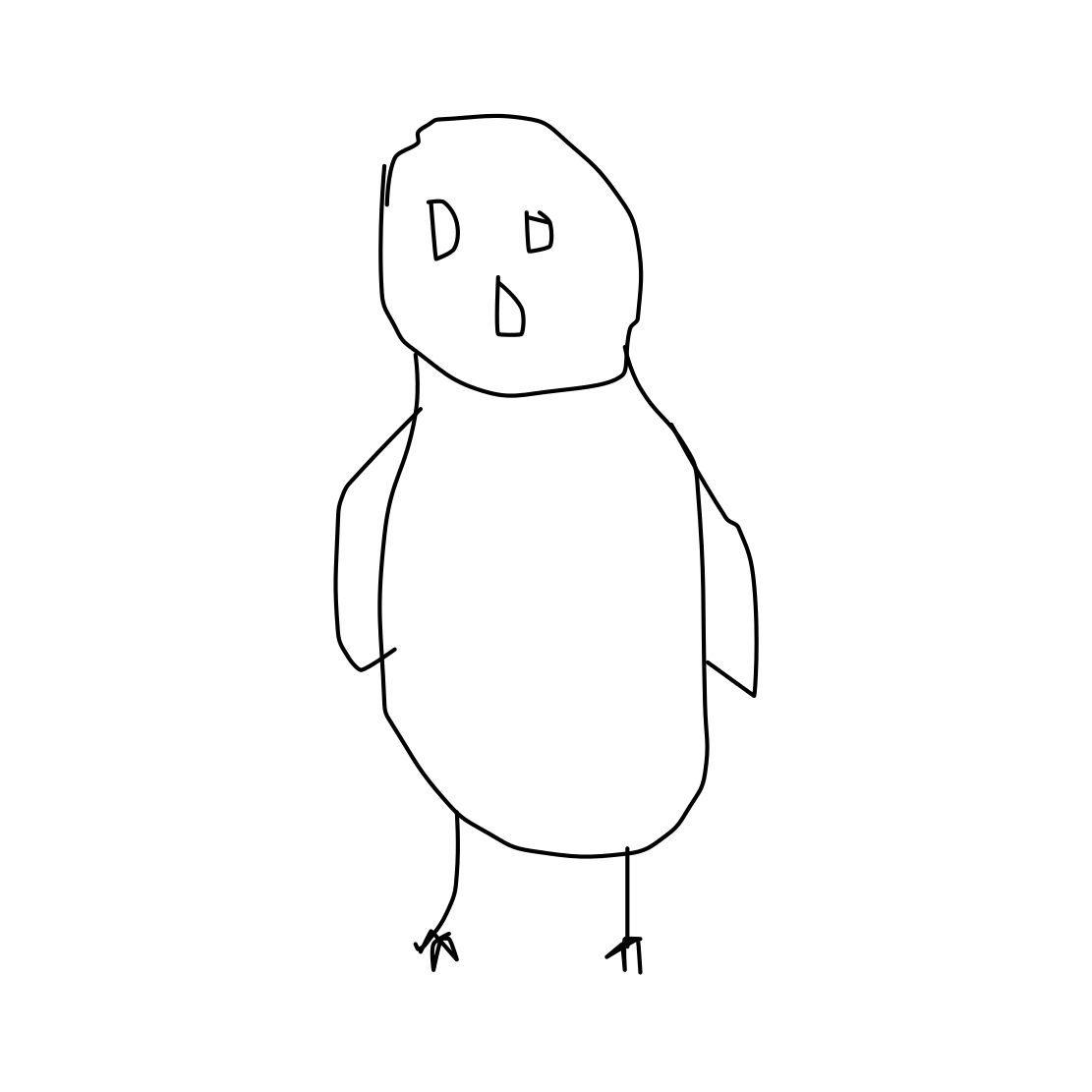} & 
\includegraphics[width=0.25\columnwidth]{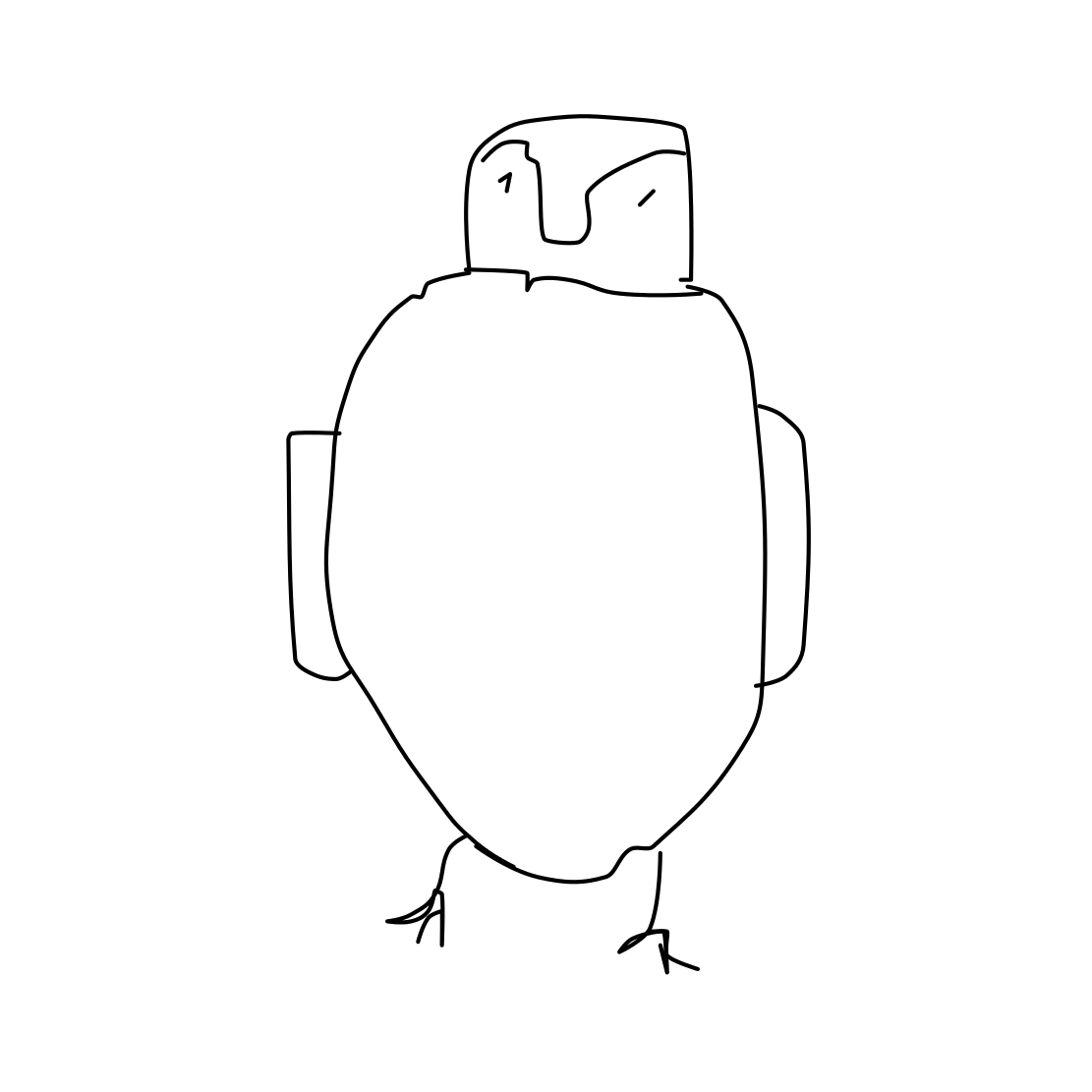} & 
\includegraphics[width=0.25\columnwidth]{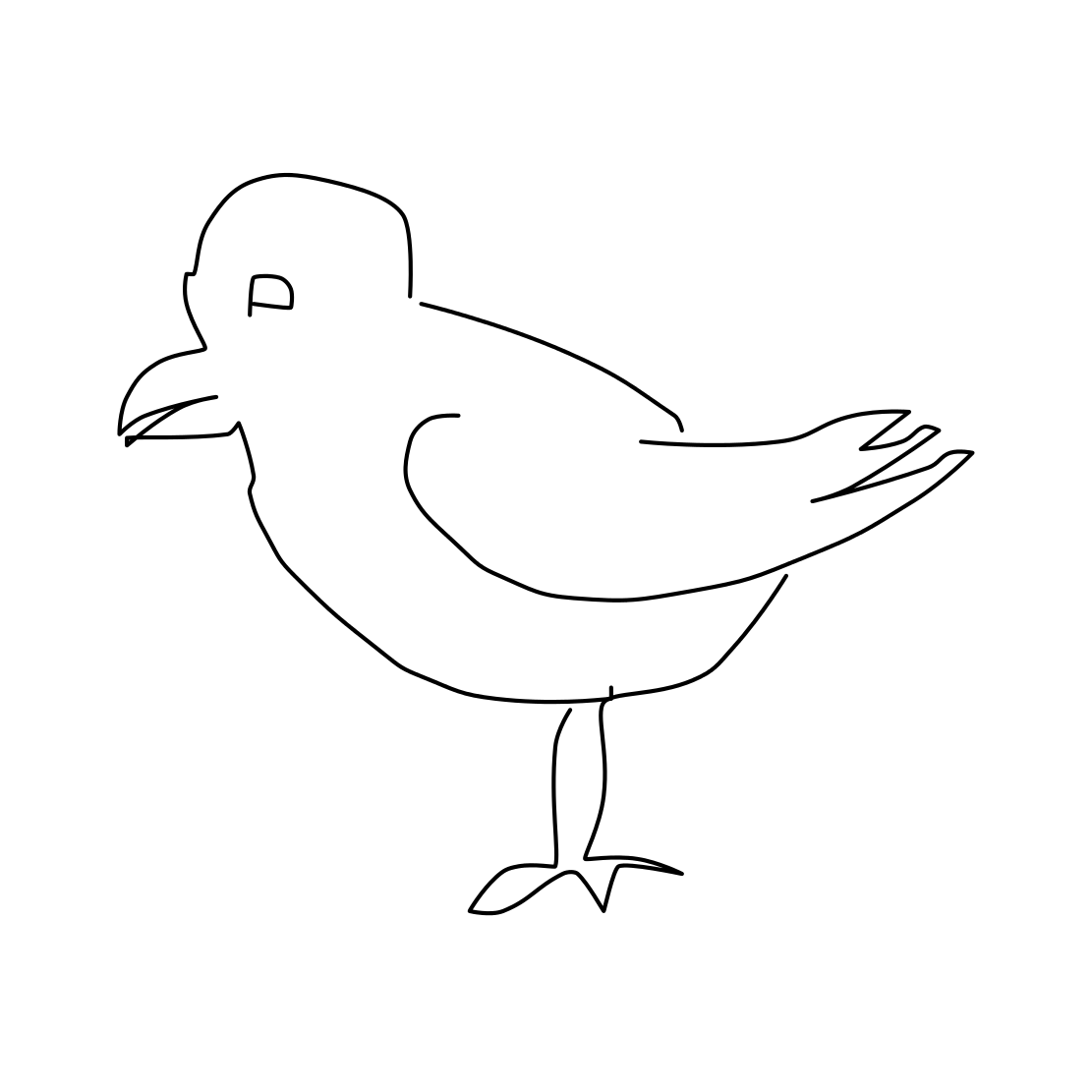} \\
\texttt{swan} & \texttt{duck} & \texttt{owl} & \texttt{penguin} & \texttt{standing bird}\\
\end{tabular}}
\caption{Inter-class similarity in TU-Berlin dataset.}
\label{tab:sketches_tu_berlin}
\end{figure}

\subsection{Comparing with the State-of-the-Art}

Apart from the two prior Zero-Shot SBIR works closest to ours, \ie ZSIH~\cite{Shen2018ZSIH} and ZS-SBIR~\cite{Yelamarthi2018ZSBIR}, we adopt fourteen ZSL and SBIR models to the zero-shot SBIR task. 
The SBIR methods that we evaluate are SaN~\cite{Yu2015}, 3D Shape~\cite{Wang2015a}, Siamese CNN~\cite{Qi2016SBIRSiamese}, GN Triplet~\cite{Sangkloy2016}, DSH~\cite{Liu2017DSH} and GDH~\cite{Zhang2018GDH}. 
A softmax baseline is also added, which is based on computing the $4096$-D VGG-$16$~\cite{Simonyan2014} feature vector pre-trained on the \emph{seen} classes for nearest neighbour search. The ZSL methods that we evaluate are: CMT~\cite{Socher2013ZSLCrossModalT}, DeViSE~\cite{Frome2013Devise}, SSE~\cite{Zhang2015ZSLSemSim}, JLSE~\cite{Zhang2016ZSLJointLatentSim}, ZSH~\cite{Yang2016ZSH}, SAE~\cite{Kodirov2017SAE} and FRWGAN~\cite{Felix2018FRWGAN}. 
We use the same \emph{seen}-\emph{unseen} splits of categories for all the experiments for a fair comparison. We compute the mean average precision (mAP@all) and precision considering top $100$ (Precision@100)~\cite{Su2015PerfEvalIR,Shen2018ZSIH} retrievals for the performance evaluation and comparison. 

\begin{figure}
\resizebox{\columnwidth}{!}{
\begin{tabular}{@{}c@{}c@{}c}
\includegraphics[width=5cm,height=5cm]{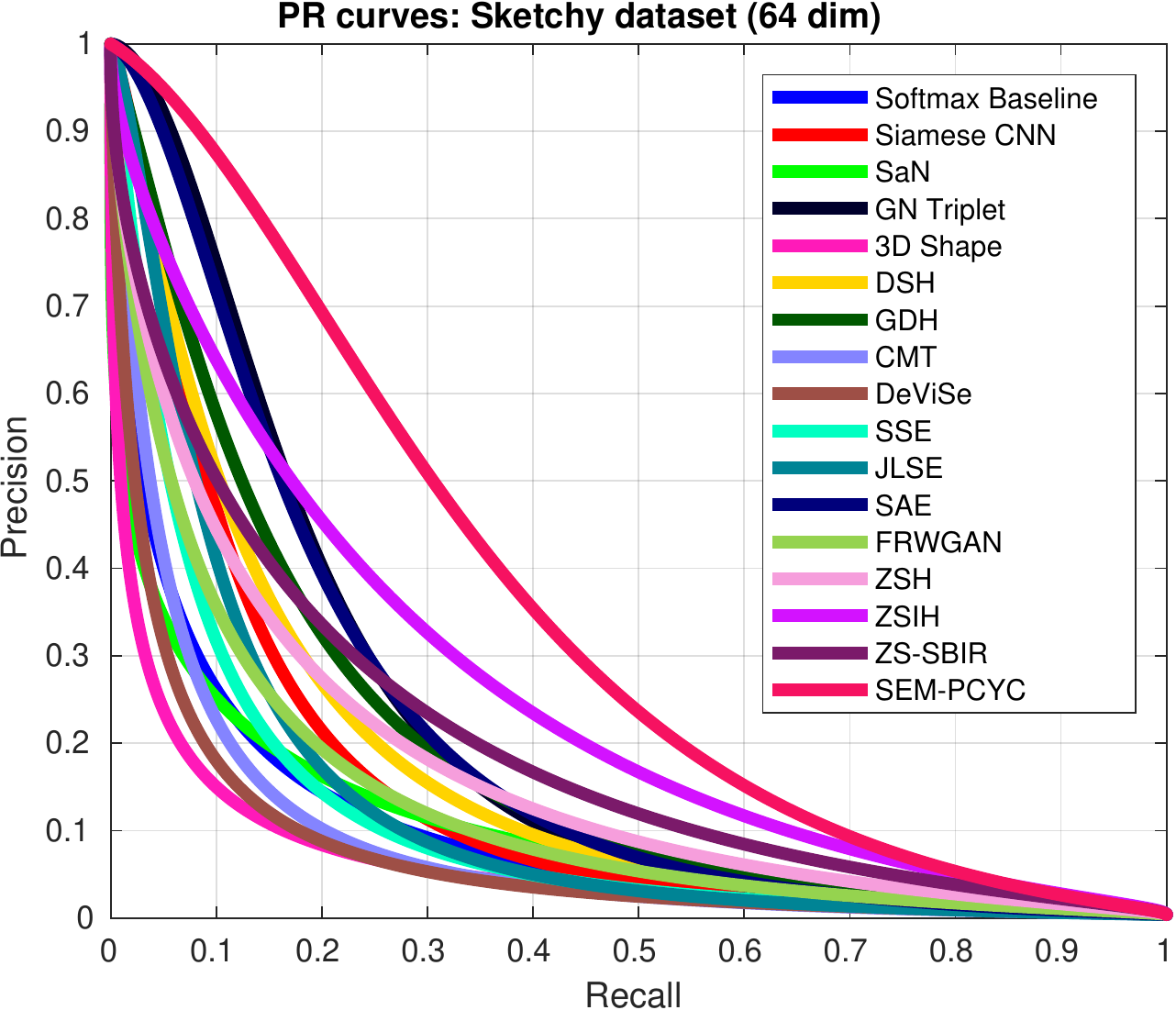} & \includegraphics[width=5cm,height=5cm]{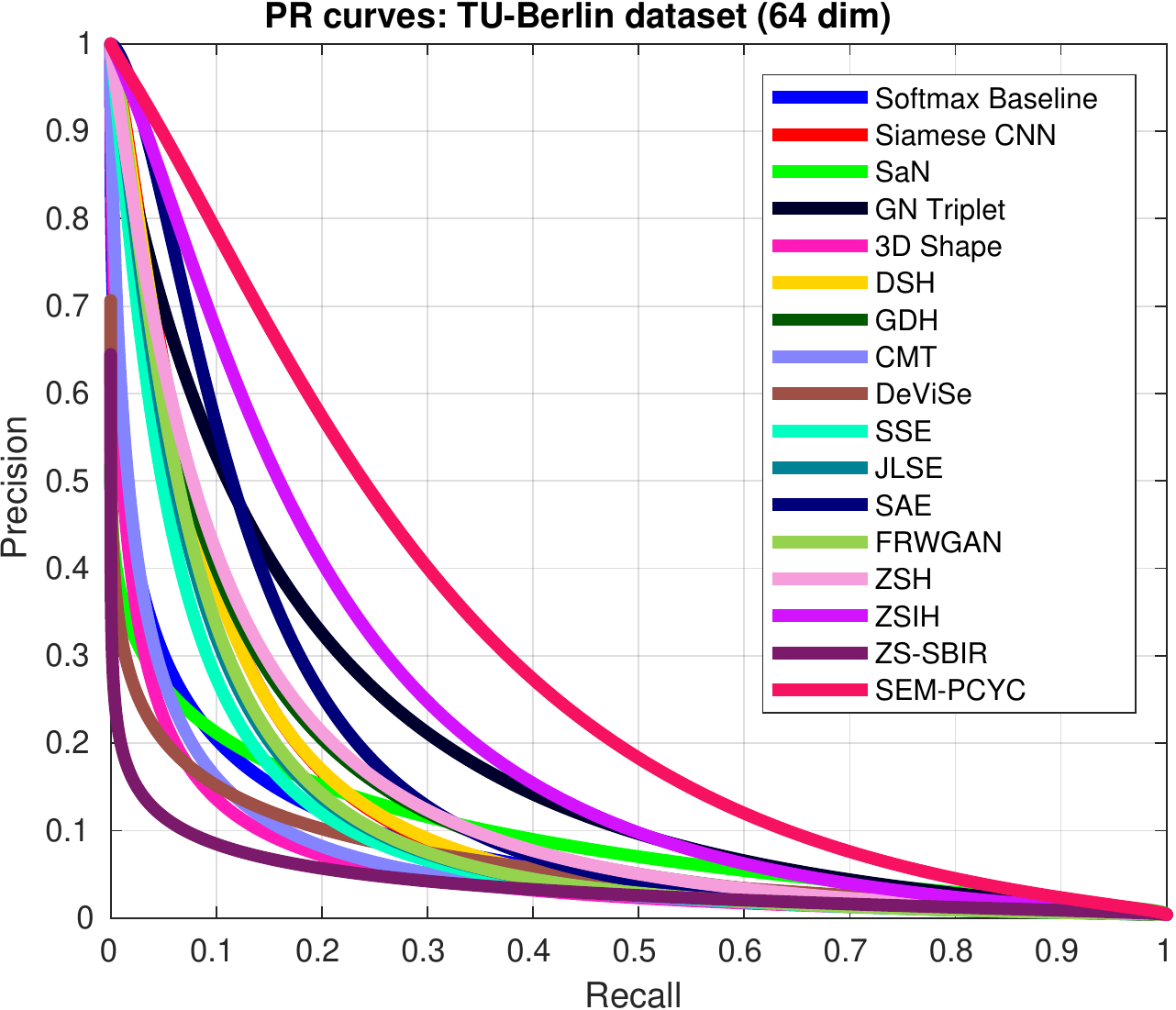} & \includegraphics[width=5cm,height=5cm]{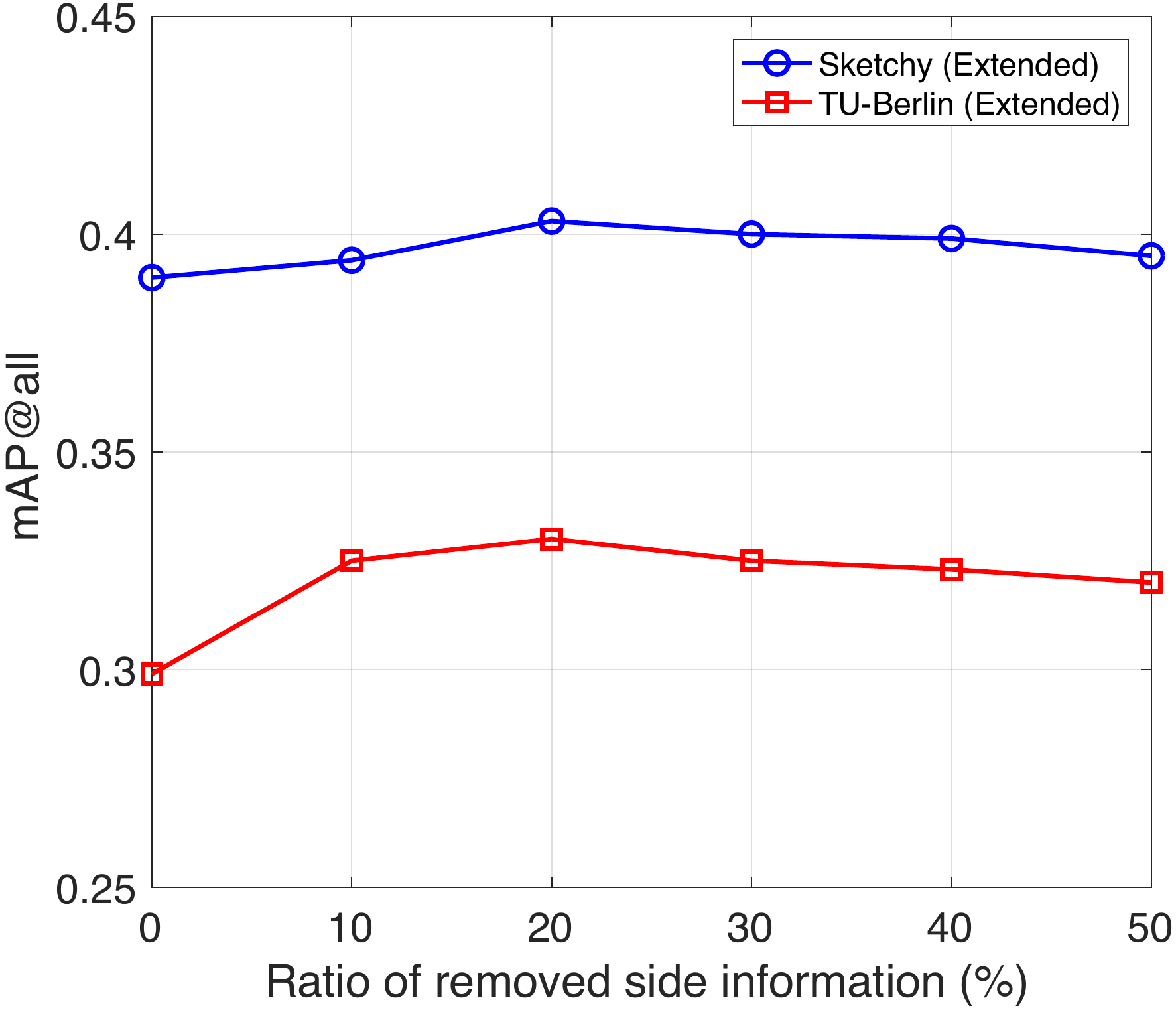}\\
(a) & (b) & (c)
\end{tabular}}
\caption{(a)-(b) PR curves of SEM-PCYC model and several SBIR, ZSL and zero-shot SBIR methods respectively on the Sketchy and TU-Berlin datasets, (c) Plot showing mAP@all wrt the ratio of removed side information. (best viewed in color)}
\label{fig:plots}
\end{figure}

Table~\ref{tab:results_z2sbir} shows that most of the SBIR and ZSL methods perform worse than the zero-shot SBIR methods. Among them, the ZSL methods usually suffer from the domain gap that exist between the sketch and image modalities while SAE~\cite{Kodirov2017SAE} reaches the best performance. The majority SBIR methods although have performed better than their ZSL counterparts, sustain the incapacity to generalize the learned representations to \emph{unseen} classes. However, GN Triplet~\cite{Sangkloy2016}, DSH~\cite{Liu2017DSH}, GDH~\cite{Zhang2018GDH} have shown reasonable potential to generalize information only from object with common shape. As per the expectation, the specialized zero-shot SBIR methods have surpassed most of the ZSL and SBIR baselines as they possess both the ability of reducing the domain gap and generalizing the learned information for the \emph{unseen} classes. ZS-SBIR learns to generalize between sketch and image from the aligned sketch-image pairs, as a result it performs well on the Sketchy dataset, but not on the TU-Berlin dataset, as in this case, aligned sketch-image pairs are not available. Our proposed method has consistently excelled the state-of-the-art method by $0.091$ mAP@all on the Sketchy dataset and $0.074$ mAP@all on the TU-Berlin dataset, which shows the effectiveness of our proposed SEM-PCYC model which gets benefited from (1) cycle consistency between sketch, image and semantic space, (2) compact and selected side information. In general, all the methods considered in \tab{tab:results_z2sbir} have performed worse on the TU-Berlin dataset, which might be due to the large number of classes, where many of them are visually similar and overlapping. These results are encouraging in that they show that the cycle consistency helps zero-shot SBIR task and our model sets the new state-of-the-art in this domain. The PR-curves of SEM-PCYC and considered baselines on Sketchy and TU-Berlin are respectively shown in~\fig{fig:plots}(a)-(b). We also conducted additional experiments on generalized ZS-SBIR setting where search space contains \emph{seen} and \emph{unseen} classes. This task is significantly more challenging than ZS-SBIR as \emph{seen} classes create distraction to the test queries. Our results in \tab{tab:results_z2sbir} (last two lines) show that our model significantly outperforms~\cite{Shen2018ZSIH}, due to the benefit of our cross-modal adversarial mechanism and heterogeneous side information. 

\input{tex/qual_results_main_small.tex} 

\myparagraph{Qualitative Results.}
Next, we analyze the retrieval performance of our proposed model qualitatively in~\fig{fig:qual_results_main} (more qualitative results are available in~\cite{Dutta2019SEM-PCYCSupp}). Some notable examples are as follows. Sketch query of \texttt{tank} retrieves some examples of \texttt{motorcycle} probably because both of them have wheels in common. For having visual and semantic similarity, sketching \texttt{guitar} retrieves some \texttt{violin}s. Querying \texttt{castle}, retrieves images having large portion of sky, because the images of its semantically similar classes, such as, \texttt{skyscraper}, \texttt{church}, are mostly captured with sky in background. In general, we observe that the wrongly retrieved candidates mostly have a closer visual and semantic relevance with the queried ones. This effect is more prominent in TU-Berlin dataset, which may be due to the inter-class similarity of sketches between different classes. As shown in \fig{tab:sketches_tu_berlin}, the classes \texttt{swan}, \texttt{duck} and \texttt{owl}, \texttt{penguin} have substantial visual similarity, and all of them are \texttt{standing bird} which is a separate class of the same dataset. Therefore, for TU-Berlin dataset, it is challenging to generalize the \emph{unseen} classes from the learned representation of \emph{seen} classes.

\subsection{Effect of Side-Information}

In zero-shot learning, side information is as important as the visual information as it is the only means the model can discover similarities between classes. As the type of side information has a high effect in performance of any method, we analyze the effect of side-information and present zero-shot SBIR results by considering different side information and their combinations. We compare the effect of using GloVe~\cite{Pennington2014GloVe} and Word2Vec~\cite{Mikolov2013a} as text-based model, and three similarity measurements, i.e. path, Lin~\cite{Lin1998ITSim} and Jiang-Conrath~\cite{Jiang1997SemSim} for constructing three different side information that are based on WordNet hierarchy. \tab{tab:res_sem} contains the quantitative results on both Sketchy and TU-Berlin datasets with different side information mentioned and their combinations, where we set $M=64$ (results with $M=32, 128$ can be found in~\cite{Dutta2019SEM-PCYCSupp}). We have observed that in majority of cases combining different side information increases the performance by $1\%$ to $3\%$. 

On Sketchy, the combination of Word2vec and Jiang-Conrath hierarchical similarity reaches the highest mAP of $0.349$ while on TU Berlin dataset, the combination of Word2Vec and path similarity leads with $0.297$ mAP. We conclude from these experiments that indeed text-based and hierarchy-based class embeddings are complementary. Furthermore, Word2Vec captures semantic similarity between words better than GloVe for the task of zero-shot SBIR.

{
\setlength{\tabcolsep}{6pt}
\renewcommand{\arraystretch}{1.0}
\begin{table}[!t]
\centering
\resizebox{\columnwidth}{!}{
\begin{tabular}{cc|ccc|c|c}
\multicolumn{2}{c|}{\textbf{Text Embedding}} & \multicolumn{3}{c|}{\textbf{Hierarchical Embedding}} & \textbf{Sketchy} & \textbf{TU-Berlin}\\
\textbf{Glove} & \textbf{Word2Vec} & \textbf{Path} & \textbf{Lin}~\cite{Lin1998ITSim} & \textbf{Ji-Cn}~\cite{Jiang1997SemSim} & \textbf{(Extended)} & \textbf{(Extended)} \\
\hline
\checkmark &  &  &  &  & $0.284$ & $0.228$ \\
 & \checkmark &  &  &  & $0.330$ & $0.232$ \\
 &  & \checkmark &  &  & $0.314$ & $0.224$ \\
 &  &  & \checkmark &  & $0.248$ & $0.169$ \\
 &  &  &  & \checkmark & $0.308$ & $0.227$ \\
 \hline
\checkmark &  & \checkmark &  &  & $0.338$ & $0.276$ \\
\checkmark &  &  & \checkmark &  & $0.299$ & $0.253$ \\
\checkmark &  &  &  & \checkmark & $0.285$ & $0.243$ \\
 & \checkmark & \checkmark &  &  & $0.340$ & $\mathbf{0.297}$ \\
 & \checkmark &  & \checkmark &  & $0.288$ & $0.264$ \\
 & \checkmark &  &  & \checkmark & $\mathbf{0.349}$ & $0.291$ \\
\end{tabular}}
\caption{Zero-shot SBIR mAP@all using different semantic embeddings (top) and their combinations (bottom).}
\label{tab:res_sem}
\end{table}
}

\subsection{Model Ablations}
\label{sec:ablation}
The baselines of our ablation study are built by modifying some parts of the SEM-PCYC model and analyze the effect of different losses of our model. First, we train the model only with adversarial loss, and then alternatively add cycle consistency and classification loss for the training. Second, we train the model without the side information selection mechanism, for that, we only take the original text or hierarchical embedding or their combination as side information, which can give an idea on the advantage of selecting side information via the auto-encoder. Next, we experiment reducing the dimensionality of the class embedding to a percentage of the full dimensionality. Finally, to demonstrate the effectiveness of the regularizer used in the auto-encoder for selecting discriminative side information, we experiment by making $\lambda=0$ in \eq{eqn:aenc_loss}.

{
\setlength{\tabcolsep}{6pt}
\renewcommand{\arraystretch}{0.9}
\begin{table}[!t]
\centering
\resizebox{\columnwidth}{!}{
\begin{tabular}{lcc}
\textbf{Description} & \textbf{Sketchy} & \textbf{TU-Berlin}\\
\hline
Only adversarial loss & $0.128$ & $0.109$ \\
Adversarial + cycle consistency loss & $0.147$ & $0.131$ \\
Adversarial + classification loss & $0.140$ & $0.127$ \\
Without selecting side information & $0.382$ & $0.299$ \\
Without regularizer in~\eq{eqn:aenc_loss} & $0.323$ & $0.273$ \\
\textbf{SEM-PCYC (full model)} & $\mathbf{0.349}$ & $\mathbf{0.297}$
\end{tabular}}
\caption{Ablation study on our $64$-D model mAP@all results of several baselines are shown above.}
\label{tab:ablation_study}
\end{table}
}

The mAP@all values obtained by respective baselines mentioned above are shown in~\tab{tab:ablation_study}. We consider the best side information setting according to~\tab{tab:res_sem} depending on the dataset. The assessed baselines have typically underperform the full SEM-PCYC model. Only with adversarial losses, the performance of our system drops significantly. We suspect that only adversarial training although maps sketch and image input to a semantic space, there is no guarantee that sketch-image pairs of same category are matched. This is because adversarial training only ensures the mapping of input modality to target modality that matches its empirical distribution~\cite{Zhu2017CycleGAN}, but does not guarantee an individual input and output are paired up. Imposition of cycle-consistency constraint ensures the one-to-one correspondence of sketch-image categories. However, the performance of our system does not improve substantially while the model is trained both with adversarial and cycle consistency loss. We speculate that this issue could be due to the lack of inter-category discriminating power of the learned embedding functions; for that, we set a classification criteria to train discriminating cross-modal embedding functions. We further observe that only imposing classification criteria together with adversarial loss, neither improves the retrieval results. We conjecture that in this case the learned embedding could be very discriminative but the two modalities might be matched in wrong way. Hence, it can be concluded that all these three losses are complimentary to each other and absolutely essential for effective zero-shot SBIR. Next, we analyze the effect of side information and observe that without the encoded and compact side information, we achieve better mAP@all with a compromise on retrieval time, as the original dimension ($354+300=654$d for Sketchy and $664+300=964$d for TU-Berlin) of considered side information is much higher than the encoded ones ($64$d). We further investigate by reducing its dimension as a percentage of the original one (see \fig{fig:plots}(c)), and we have observed that at the beginning, reducing a small part (mostly $5\%$ to $30\%$) usually leads to a better performance, which reveals that not all the side information are necessary for effective zero-shot SBIR and some of them are even harmful. In fact, the first removed ones have low information content, and can be regarded as noise. We have also perceived that removing more side information (beyond $20\%$ to $40\%$) deteriorates the performance of the system, which is quite justifiable because the compressing mechanism of auto-encoder progressively removes important and predictable side information. However, it can be observed that with highly compressed side information as well, our model provides a very good deal with performance and retrieval time. Without using the regularizer in \eq{eqn:aenc_loss}, although our system performs reasonably, the mAP@all value is still lower than the best obtained performance. We explain this as a benefit of using $\ell_{21}$-norm based regularizer that effectively select representative side information.


%% file: tex/qual_results_main_small.tex
\begin{figure}
\begin{center}
\resizebox{\columnwidth}{!}{
\begin{tabular}{@{}c@{}c@{}c@{}c@{}c@{}c@{}c@{}c@{}c@{}c@{}c}
\includegraphics[width=1.5cm, height=1.5cm]{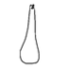} & \includegraphics[width=1.5cm, height=1.5cm]{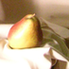} & \includegraphics[width=1.5cm, height=1.5cm]{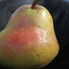} & \includegraphics[width=1.5cm, height=1.5cm]{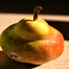} & \includegraphics[width=1.5cm, height=1.5cm]{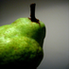} & \includegraphics[width=1.5cm, height=1.5cm]{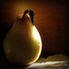} & \includegraphics[width=1.5cm, height=1.5cm]{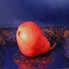} & \includegraphics[width=1.5cm, height=1.5cm]{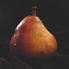} & \includegraphics[width=1.5cm, height=1.5cm]{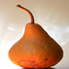} & \includegraphics[width=1.5cm, height=1.5cm]{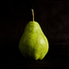} & \includegraphics[width=1.5cm, height=1.5cm]{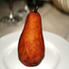} \\
 & \cmark & \cmark & \cmark & \cmark & \cmark & \cmark & \cmark & \cmark & \cmark & \cmark \\
 \includegraphics[width=1.5cm, height=1.5cm]{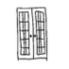} & \includegraphics[width=1.5cm, height=1.5cm]{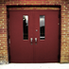} & \includegraphics[width=1.5cm, height=1.5cm]{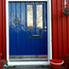} & \includegraphics[width=1.5cm, height=1.5cm]{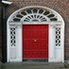} & \includegraphics[width=1.5cm, height=1.5cm]{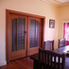} & \includegraphics[width=1.5cm, height=1.5cm]{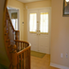} & \includegraphics[width=1.5cm, height=1.5cm]{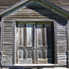} & \includegraphics[width=1.5cm, height=1.5cm]{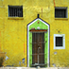} & \includegraphics[width=1.5cm, height=1.5cm]{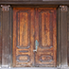} & \includegraphics[width=1.5cm, height=1.5cm]{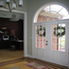} & \includegraphics[width=1.5cm, height=1.5cm]{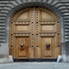} \\
 & \cmark & \cmark & \cmark & \cmark & \cmark & \cmark & \cmark & \cmark & \cmark & \cmark \\
\includegraphics[width=1.5cm, height=1.5cm]{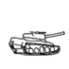} & \includegraphics[width=1.5cm, height=1.5cm]{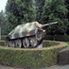} & \includegraphics[width=1.5cm, height=1.5cm]{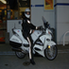} & \includegraphics[width=1.5cm, height=1.5cm]{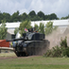} & \includegraphics[width=1.5cm, height=1.5cm]{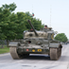} & \includegraphics[width=1.5cm, height=1.5cm]{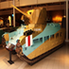} & \includegraphics[width=1.5cm, height=1.5cm]{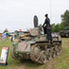} & \includegraphics[width=1.5cm, height=1.5cm]{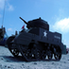} & \includegraphics[width=1.5cm, height=1.5cm]{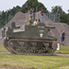} & \includegraphics[width=1.5cm, height=1.5cm]{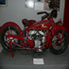} & \includegraphics[width=1.5cm, height=1.5cm]{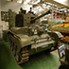} \\
 & \cmark & \xmark & \cmark & \cmark & \cmark & \cmark & \cmark & \cmark & \xmark & \cmark \\
\includegraphics[width=1.5cm, height=1.5cm]{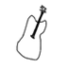} & \includegraphics[width=1.5cm, height=1.5cm]{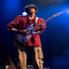} & \includegraphics[width=1.5cm, height=1.5cm]{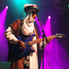} & \includegraphics[width=1.5cm, height=1.5cm]{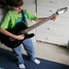} & \includegraphics[width=1.5cm, height=1.5cm]{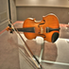} & \includegraphics[width=1.5cm, height=1.5cm]{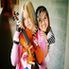} & \includegraphics[width=1.5cm, height=1.5cm]{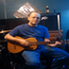} & \includegraphics[width=1.5cm, height=1.5cm]{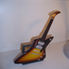} & \includegraphics[width=1.5cm, height=1.5cm]{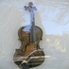} & \includegraphics[width=1.5cm, height=1.5cm]{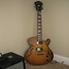} & \includegraphics[width=1.5cm, height=1.5cm]{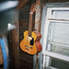} \\
 & \cmark & \cmark & \cmark & \xmark & \xmark & \cmark & \cmark & \xmark & \cmark & \cmark \\
\includegraphics[width=1.5cm, height=1.5cm]{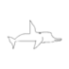} & \includegraphics[width=1.5cm, height=1.5cm]{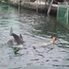} & \includegraphics[width=1.5cm, height=1.5cm]{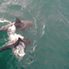} & \includegraphics[width=1.5cm, height=1.5cm]{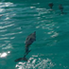} & \includegraphics[width=1.5cm, height=1.5cm]{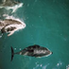} & \includegraphics[width=1.5cm, height=1.5cm]{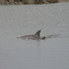} & \includegraphics[width=1.5cm, height=1.5cm]{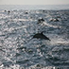} & \includegraphics[width=1.5cm, height=1.5cm]{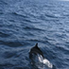} & \includegraphics[width=1.5cm, height=1.5cm]{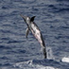} & \includegraphics[width=1.5cm, height=1.5cm]{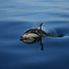} & \includegraphics[width=1.5cm, height=1.5cm]{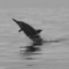} \\ 
 & \cmark & \cmark & \cmark & \cmark & \cmark & \cmark & \cmark & \cmark & \cmark & \cmark \\
\includegraphics[width=1.5cm, height=1.5cm]{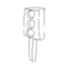} & \includegraphics[width=1.5cm, height=1.5cm]{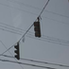} & \includegraphics[width=1.5cm, height=1.5cm]{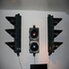} & \includegraphics[width=1.5cm, height=1.5cm]{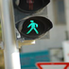} & \includegraphics[width=1.5cm, height=1.5cm]{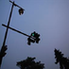} & \includegraphics[width=1.5cm, height=1.5cm]{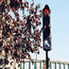} & \includegraphics[width=1.5cm, height=1.5cm]{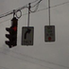} & \includegraphics[width=1.5cm, height=1.5cm]{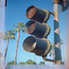} & \includegraphics[width=1.5cm, height=1.5cm]{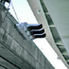} & \includegraphics[width=1.5cm, height=1.5cm]{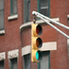} & \includegraphics[width=1.5cm, height=1.5cm]{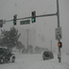} \\ 
 & \cmark & \cmark & \cmark & \cmark & \cmark & \cmark & \cmark & \cmark & \cmark & \cmark \\
\includegraphics[width=1.5cm, height=1.5cm]{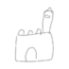} & \includegraphics[width=1.5cm, height=1.5cm]{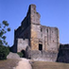} & \includegraphics[width=1.5cm, height=1.5cm]{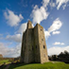} & \includegraphics[width=1.5cm, height=1.5cm]{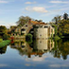} & \includegraphics[width=1.5cm, height=1.5cm]{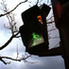} & \includegraphics[width=1.5cm, height=1.5cm]{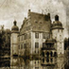} & \includegraphics[width=1.5cm, height=1.5cm]{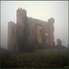} & \includegraphics[width=1.5cm, height=1.5cm]{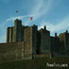} & \includegraphics[width=1.5cm, height=1.5cm]{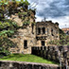} & \includegraphics[width=1.5cm, height=1.5cm]{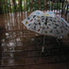} & \includegraphics[width=1.5cm, height=1.5cm]{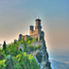} \\ 
 & \cmark & \cmark & \cmark & \xmark & \cmark & \cmark & \cmark & \cmark & \xmark & \cmark \\
\includegraphics[width=1.5cm, height=1.5cm]{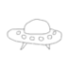} & \includegraphics[width=1.5cm, height=1.5cm]{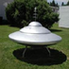} & \includegraphics[width=1.5cm, height=1.5cm]{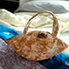} & \includegraphics[width=1.5cm, height=1.5cm]{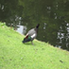} & \includegraphics[width=1.5cm, height=1.5cm]{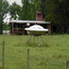} & \includegraphics[width=1.5cm, height=1.5cm]{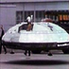} & \includegraphics[width=1.5cm, height=1.5cm]{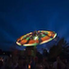} & \includegraphics[width=1.5cm, height=1.5cm]{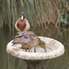} & \includegraphics[width=1.5cm, height=1.5cm]{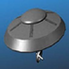} & \includegraphics[width=1.5cm, height=1.5cm]{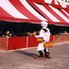} & \includegraphics[width=1.5cm, height=1.5cm]{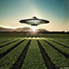} \\ 
 & \cmark & \xmark & \xmark & \cmark & \cmark & \cmark & \xmark & \cmark & \xmark & \cmark \\ 
\end{tabular}
}
\end{center}
\caption{Top-10 zero-shot SBIR results obtained by our SEM-PCYC model on Sketchy (top four rows) and TU-Berlin (next four rows) are shown here according to the Euclidean distances, where the green ticks denote correctly retrieved candidates and the red crosses indicate wrong retrievals. (best viewed in color)}
\label{fig:qual_results_main}
\end{figure}

%% file: tex/concl.tex
\section{Conclusion}
\label{sec:concl}
We proposed the SEM-PCYC model for the zero-shot SBIR task. Our SEM-PCYC is a semantically aligned paired cycle consistent generative model whose each branch either maps a sketch or an image to a common semantic space via adversarial training with a shared discriminator. Thanks to cycle consistency on both the branches our model does not require aligned sketch-image pairs. Moreover, it acts as a regularizer in the adversarial training. The classification losses on the generators guarantee the features to be discriminative. We show that combining heterogeneous side information through an auto-encoder, which encodes a compact side information useful for adversarial training, is effective. Our evaluation on two datasets has shown that our model consistently outperforms the existing methods in zero-shot SBIR task.

%% file: tex/ack.tex
\section*{Acknowledgments}
This work has been partially supported by European Union's research and innovation program under Marie Sk\l{}odowska-Curie grant agreement No. 665919. The Titan Xp and Titan V used for this research were donated by the NVIDIA Corporation.